\documentclass{article}

\usepackage[preprint]{neurips_2026}

\usepackage[utf8]{inputenc} 
\usepackage[T1]{fontenc}    
\usepackage[hidelinks]{hyperref}       
\usepackage{url}            
\usepackage{booktabs}       
\usepackage[acronym, nonumberlist]{glossaries}     
\usepackage{amsfonts}       
\usepackage{nicefrac}       
\usepackage{microtype}      
\usepackage{xcolor}         
\usepackage{amsmath}
\usepackage{amssymb}
\usepackage{mathtools}
\usepackage{amsthm}
\usepackage{comment}
\usepackage[normalem]{ulem}
\usepackage{dsfont}
\usepackage{tcolorbox}
\usepackage{xcolor}
\usepackage{algorithm}
\usepackage{algorithmic}
\newtcolorbox{algobox}{
  colback=white,
  colframe=black,
  boxrule=0.3pt,
  arc=3pt,
  left=2pt,
  right=2pt,
  top=0pt,
  bottom=0pt,
  before skip=2pt,
  after skip=4pt,
  enlarge left by=-5pt,
}

\theoremstyle{plain}

\theoremstyle{definition}

\theoremstyle{remark}

\newacronym{acr:rl}{RL}{Reinforcement Learning}
\newacronym{acr:dqn}{DQN}{Deep Q-Network}
\newacronym{acr:per}{PER}{Prioritized Experience Replay}
\newacronym{acr:reaper}{ReaPER}{Reliability-adjusted Prioritized Experience Replay}
\newacronym{acr:sac}{SAC}{Soft Actor-Critic}
\newacronym{acr:ddqn}{DDQN}{Double Deep Q-Network}
\newacronym{acr:tql}{TQL}{Tabular Q-Learning}
\newacronym{acr:td}{TD}{Temporal Difference}
\newacronym{acr:off_tde}{Off-TDE}{Offline Temporal Difference Error}
\newacronym{acr:on_tde}{On-TDE}{Online Temporal Difference Error}
\newacronym{acr:mdp}{MDP}{Markov Decision Process}
\newacronym{acr:tarl}{TARL}{Target-Aligned Reinforcement Learning}
\newacronym{acr:tadqn}{TA-DQN}{Target-Aligned DQN}
\newacronym{acr:taddqn}{TA-DDQN}{Target-Aligned DDQN}
\newacronym{acr:tasac}{TA-SAC}{Target-Aligned SAC}
\newacronym{acr:nauc}{nAUC}{normalized Area Under the Curve}
\makeglossaries
\glsdisablehyper

\title{Target-Aligned Reinforcement Learning}

\author{%
  Leonard S.~Pleiss\\
  Technical University of Munich\\
  \texttt{leonard.pleiss@tum.de} \\
  \AND
  James Harrison \\
  Google DeepMind \\
  \texttt{jamesharrison@google.com} \\
  \And
  Maximilian Schiffer \\
  HEC Paris\\
  \texttt{schiffer@hec.fr} \\
}

\begin{document}
\newcommand{\idxCurrent}{c}
\newcommand{\idxPrimary}{t}
\newcommand{\idxSecondary}{i}
\newcommand{\idxTertiary}{m}
\newcommand{\idxOracle}{l}
\newcommand{\idxSelected}{j}

\maketitle

\begin{abstract}
    Many value-based deep reinforcement learning algorithms rely on target networks---lagged copies of the online network---to stabilize training. While effective, this mechanism introduces a fundamental stability-recency trade-off: slower target updates improve stability but reduce the recency of learning signals, hindering convergence speed. We propose Target-Aligned Reinforcement Learning (TARL), a simple drop-in refinement for existing algorithms that emphasizes transitions for which the target and online network estimates are highly aligned. By focusing updates on well-aligned targets, TARL mitigates the adverse effects of stale target estimates while retaining the stabilizing benefits of target networks. We empirically demonstrate consistent improvements within discrete and continuous control algorithms across various benchmark environments without any hyperparameter tuning, including a 38.18\% median peak score gain on Atari-10, while incurring less than a 4\% increase in wall-clock time.
\end{abstract}

\section{Introduction}

In \gls{acr:rl}, agents optimize their policy through repeated interaction with an environment, aiming to maximize expected cumulative return. A central mechanism enabling this process is \gls{acr:td} learning \citep{sutton1988temporaldifference}, which relies on bootstrapped target estimates---so-called target values. Current state estimates are iteratively updated toward the estimates of future states, thereby allowing the agent to anticipate long-term reward.

However, as shown by \citet{mnih_human-level_2015}, naïvely using the online network to produce target values can destabilize learning. To address this issue, they introduced a designated target network. A target network is a lagged copy of the online network used exclusively to generate target values. By decoupling the online parameter updates from target value computation, the bootstrapped targets are generated by a comparatively stationary function approximator. This reduces temporal correlations between updates and targets, suppresses self-reinforcing error feedback, and substantially improves training stability and convergence behavior. This innovation contributed to landmark achievements such as human-level performance on the Atari benchmark \citep{mnih_human-level_2015}, and remains widely used \citep{fujimoto2018addressingfunctionapproximationerror, hessel_rainbow_2017, haarnoja_soft_2018, badia_agent57_2020, schwarzer_bigger_2023}.

Despite their widespread success, target networks introduce a critical limitation: by construction, their estimates are lagged relative to the online network. The lag can cause target estimates to become stale, reflecting outdated value predictions that no longer align with the current policy or representation \citep{vincent2025bridgingperformancegaptargetfree}. In essence, it creates a dynamic analogous to aiming at a moving target using delayed coordinates: if the value function changes direction, the lagged target steers the update astray. Consequently, improved target stability necessarily comes at the cost of reduced recency, while increasing recency risks reintroducing instability. This phenomenon constitutes the fundamental \emph{stability-recency trade-off}: Ideally, target values should be as up-to-date as possible while preserving the level of stability required for reliable bootstrapped learning.

In this work, we argue that sacrificing recency is not necessary to ensure stability. We propose to mitigate this trade-off by using the unstable but recent online network to validate the stable but partially stale target network. Our approach---\gls{acr:tarl}---prioritizes updates where online and target estimates agree, enabling the agent to exploit recent information without compromising stability.

\textbf{Background.} Even though individual papers have recently claimed to bypass the need for a target network by proposing alternate update rules \citep{ijcai2019p379}, functional regularization \citep{piche2023bridginggaptargetnetworks}, decision transformers \citep{chen2021decisiontransformerreinforcementlearning} or streaming algorithms \citep{elsayed2024deep}, target networks remain the de facto standard in value-based \gls{acr:rl}.
They are a core component in most popular \gls{acr:rl} algorithms, including \gls{acr:dqn}, Rainbow, \gls{acr:sac} or Deep Deterministic Policy Gradient \citep{lillicrap_continuous_2015, hessel_rainbow_2017, mnih_human-level_2015, haarnoja_soft_2018}. Target networks further remain a cornerstone of state-of-the-art \gls{acr:rl} algorithms to this day \citep{badia_agent57_2020, hessel_rainbow_2017, haarnoja_soft_2018, schwarzer_bigger_2023}.

As such, the stability-recency trade-off remains a fundamental consideration in \gls{acr:rl}, particularly during hyperparameter tuning. Most algorithms navigate the trade-off through design choices that govern the target network's update mechanism and cadence. There are two major update mechanisms, \emph{hard} and \emph{soft} target network updates \citep{Li2023}.

Methods like \gls{acr:dqn} \citep{mnih_human-level_2015} traditionally perform \emph{hard} updates. At some fixed interval---denoted as the target update interval $K$---the target network is overwritten by the online network. This interval is then tuned to balance stability and recency.

Other algorithms like \gls{acr:sac} or Deep Deterministic Policy Gradient usually perform \emph{soft} updates \citep{lillicrap_continuous_2015, KOBAYASHI202163, haarnoja_soft_2018, pmlr-v139-zhang21y}. A soft update is a mechanism to incrementally adjust the target network's parameters toward those of the online network. Rather than copying parameters at fixed intervals, the target network is gradually updated at every step. These gradual updates are usually performed via exponential moving or Polyak averages \citep{polyak}, typically of the form

\vspace{-1em}
\begin{equation}
\theta_{\text{target}} \leftarrow \tau \theta_{\text{online}} + (1 - \tau)\theta_{\text{target}},
\end{equation}
\vspace{-1em}

where $\tau \in (0,1]$ controls the update magnitude. Soft updates reduce the variance and non-stationarity of bootstrapped targets by smoothing parameter changes over time, improving training stability while still allowing target estimates to track the evolving online network.

Regardless of whether frequent soft updates or infrequent hard updates are employed, the challenge remains the same: to identify an optimal balance where stability is maintained while recency is maximized. Current methods therefore essentially treat this trade-off as a zero-sum game, merely adjusting the slider between stability (low $\tau$ or high $K$) and recency (high $\tau$ or low $K$).

In contrast, we approach this challenge from a novel angle. We posit that we do not need to make the target network fresher (and more unstable); we only need to defer updates where the target network's staleness currently leads to outdated update directions. Specifically, we focus on the alignment between the stable estimate---the target network's target value---and the most recent estimate---the online network's target value. We propose to prioritize updates where both target estimates are well-aligned, indicating that our learning target is both stable \emph{and} recent. In doing so, \gls{acr:tarl} is the first method to address the stability-recency trade-off inherent to target networks on a per-sample basis, shifting from network-level approaches (e.g., tuning update frequency or Polyak coefficients) to individual transitions. Operating at this finer granularity bypasses the conventional trade-off, preserving stability without sacrificing recency.

\textbf{Contribution.} Our contribution is threefold: First, we introduce a novel offline-online target alignment metric that quantifies agreement between the online and target network value estimates. Second, we present \gls{acr:tarl}, an algorithmic framework that integrates target alignment through oversampling into any standard \gls{acr:rl} algorithm employing target networks. Finally, we support our claims with empirical studies across various discrete and continuous control benchmarks and multiple algorithms, showing that \gls{acr:tarl} consistently improves performance.

\section{Problem statement}\label{sec:problem_statement}

We consider a standard \gls{acr:mdp} as usually studied in an \gls{acr:rl} setting \citep{sutton_reinforcement_1998}. We characterize this \gls{acr:mdp} as a tuple $\left(\mathcal{S}, \mathcal{A}, P, r, \gamma, p\right)$, where $\mathcal{S}$ is a state space, $\mathcal{A}$ is an action space, $P: \mathcal{S} \times \mathcal{A} \to \Delta(\mathcal{S})$ is a stochastic kernel, $r: \mathcal{S} \times \mathcal{A} \to \mathbb{R}$ is a reward function, $\gamma\in(0, 1)$ is a discount factor, and $p\in\Delta(\mathcal{S})$ denotes a probability mass function describing the distribution of the initial state, $S_1\sim p$.

We denote by $S_{\idxPrimary}$ and $A_{\idxPrimary}$ the random variables representing the state and action at time $\idxPrimary$, and by $s \in \mathcal{S}$ and $a \in \mathcal{A}$ their respective realizations. At time step $\idxPrimary$, the system is in state $S_{\idxPrimary}\in\mathcal{S}$. If an agent takes action $A_{\idxPrimary}\in\mathcal{A}$, it receives a corresponding reward $r(s,a)$, and the system transitions to the next state $S_{\idxPrimary+1}\sim P(\cdot|s,a)$. We define the random reward at time $\idxPrimary$ as $R_{\idxPrimary} = r(S_{\idxPrimary},A_{\idxPrimary})$. The agent selects actions based on a policy $\pi:\mathcal{S}\to\mathcal{A}$ via $A_{\idxPrimary} = \pi(S_{\idxPrimary})$. Let $\mathbb{P}_{p}^{\pi}(\cdot)=\textup{Prob}(\cdot\mid\pi,S_1\sim p)$ denote the probability of an event when following a policy $\pi$, starting from an initial state $S_1\sim p$, and let $\mathbb{E}_p^\pi[\cdot]$ denote the corresponding expectation operator. Let $n$ denote the number of transitions within the episode. Let $G_{\idxPrimary}$ denote the discounted return at time $\idxPrimary$, with $G_{\idxPrimary} = \sum_{\idxSecondary=\idxPrimary}^n \gamma^{\idxSecondary-t}R_{\idxSecondary}$. We define the Q-function (or action-value function) for a policy $\pi$ as

\vspace{-1em}
\begin{align} \label{eq:qfunction}
Q^\pi(s, a) = \mathbb{E}_p^\pi\left[G_{\idxPrimary} \mid S_{\idxPrimary}=s, A_{\idxPrimary}=a\right] = \mathbb{E}_p^\pi\left[\sum_{\idxSecondary=\idxPrimary}^n \gamma^{\idxSecondary-\idxPrimary}R_{\idxSecondary} \mid S_{\idxPrimary}=s, A_{\idxPrimary}=a\right],
\end{align}
\vspace{-1em}

where action $a$ is taken at time $t$ and $\pi$ is followed thereafter. The ultimate goal of value-based \gls{acr:rl} is to learn a policy that maximizes the Q-function, leading to $Q^\star(s,a) = \max_\pi Q^\pi(s,a)$. 

The policy is gradually improved by repeatedly interacting with the environment and learning from previously experienced transitions. A transition $C_{\idxPrimary}$ is a 5-tuple, $C_{\idxPrimary} = (S_{\idxPrimary}, A_{\idxPrimary}, R_{\idxPrimary}, S_{\idxPrimary+1}, d_{\idxPrimary})$, where $d_{\idxPrimary}$ is a binary episode termination indicator, $d_{\idxPrimary} = \mathds{1}_{\idxPrimary = n}$. One popular approach to learn $Q^\star$ is via Watkins' Q-learning \citep{watkins_learning_1989, watkins_q-learning_1992}, where Q-values are gradually updated via

\vspace{-1em}
\begin{equation} \label{eq:proposed:update}
Q(S_{\idxPrimary}, A_{\idxPrimary}) \gets Q(S_{\idxPrimary}, A_{\idxPrimary}) + \eta \cdot \delta_{\idxPrimary},
\end{equation}
\vspace{-1em}

with $\eta \in (0,1]$ being the learning rate and $\delta_{\idxPrimary}$ being the \gls{acr:td} error. In the classical tabular formulation, this \gls{acr:td} error is defined using the \emph{online target},

\vspace{-1em}
\begin{equation} \label{eq:on_target}
Q_{\text{target}}(S_{\idxPrimary}) = R_{\idxPrimary} + (1 - d_{\idxPrimary}) \cdot \gamma \cdot \max_a Q(S_{\idxPrimary+1}, a, \theta),
\end{equation}
\vspace{-1em}

yielding the online \gls{acr:td} error $\delta_{\idxPrimary} = Q_{\text{target}}(S_{\idxPrimary}) - Q(S_{\idxPrimary}, A_{\idxPrimary}, \theta)$.

In deep \gls{acr:rl}, and specifically in the \gls{acr:dqn} algorithm \citep{mnih_human-level_2015}, stability is improved by introducing a separate \emph{target network} with parameters $\Bar{\theta}$. Crucially, $\Bar{\theta}$ is a time-lagged copy of the online parameters $\theta$, such that, under hard updates, $\Bar{\theta} \approx \theta_{t-k}$ for some delay $k$. This target network improves training stability: When a single function approximator is used to both estimate action values and define the bootstrap targets, parameter updates induce correlated changes in predictions and targets, which can lead to divergence. The target network mitigates this issue by decoupling target computation from the online updates, yielding a temporally more stationary learning objective.

The \emph{offline target}---i.e., the target value from the target network, as used in standard \gls{acr:dqn}---is defined as 

\vspace{-1em}
\begin{equation} \Bar{Q}_{\text{target}}(S_{t}) = R_{t} + (1 - d_{t}) \cdot \gamma \cdot \max_a Q(S_{t+1}, a, \Bar{\theta}),
\end{equation}
\vspace{-1em}

which leads to the offline or target \gls{acr:td} error $\bar{\delta}_t = \Bar{Q}_{\text{target}}(S_{t}) - Q(S_{t}, A_{t},\theta)$.

\section{Target-Aligned Reinforcement Learning}\label{sec:methodology}

In this section, we present \gls{acr:tarl}, a framework designed to mitigate the stability-recency trade-off in deep \gls{acr:rl}. We first build the intuition for target alignment by analyzing the interplay between offline and online value estimates. We then formalize this intuition into a computable alignment score. Finally, we introduce an alignment-based oversampling mechanism that integrates this metric into standard off-policy algorithms to prioritize stable \emph{and} recent updates.

\subsection{Target alignment}\label{sec:intuition}
\begin{figure}[t!]
    \centering
    \vspace{-2.5em}
    \includegraphics[width=1\linewidth]{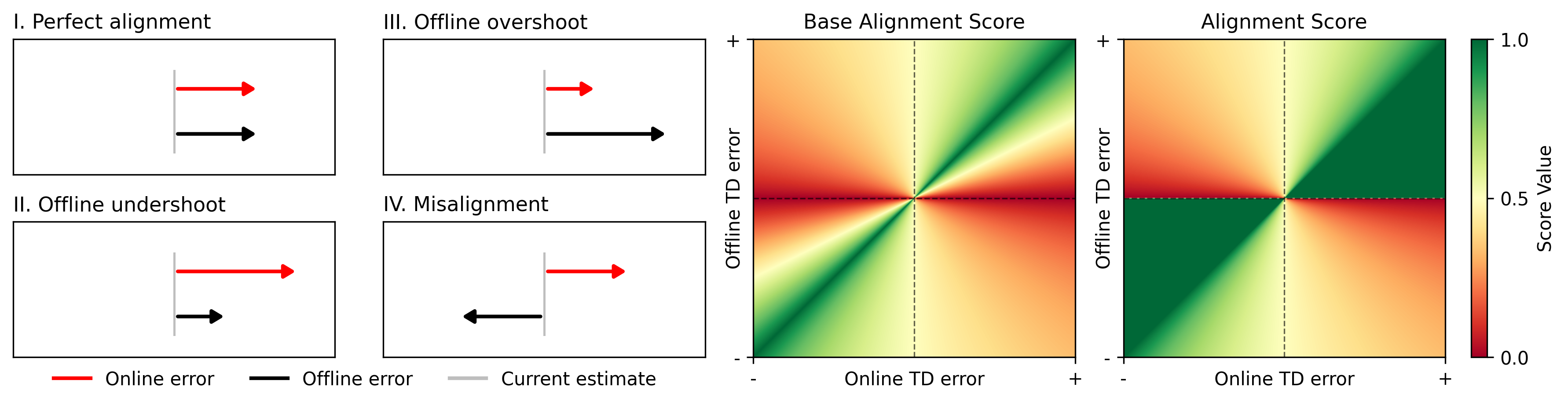}
    \vspace{-2em}
    \caption{
    \emph{Left:} Stylized visualization of alignment scenarios. We distinguish between updates fully supported by the online network (I and II) and updates that are only partially or not at all supported by the online network (III and IV). \emph{Right:} Visualization of base alignment and alignment as a function of online and offline temporal difference error.
    }
    \vspace{-0.4cm}
    \label{fig:alignment_intuition}
\end{figure}

In contemporary deep \gls{acr:rl}, value estimates are updated using target values provided by a lagged offline network. While this lag increases stability, it naturally sacrifices recency. Because the target network lacks the most recent updates, it is prone to staleness bias. \gls{acr:tarl} targets this staleness bias by quantifying the agreement between the stable (offline) target and the most recent (online) estimate. Specifically, it measures the extent to which the update direction and magnitude proposed by the offline target are \textit{supported} by the online network. As displayed in Figure~\ref{fig:alignment_intuition}, we identify four distinct alignment scenarios.

\noindent \emph{I) Perfect Alignment.} The offline and online updates are identical. The offline target is perfectly aligned with the most recent estimate from the online network.

\emph{II) Offline Undershoot.} Both networks agree on direction, but the online network proposes a more extreme update. The conservative offline update is therefore \textit{fully supported} by the online view. We treat this as a safe update: the stable target points in the right direction but is simply more cautious than the volatile online network.

\emph{III) Offline Overshoot.} Both networks agree on direction, but the offline target proposes a more extreme update than the online network. The offline update is therefore only \textit{partially supported}, indicating some level of risk that the frozen offline target overshoots the ideal update magnitude, introducing the potential need for partial update reversion in the future. In essence, the stale target may push the agent too hard, ignoring recent evidence that the value is actually closer than the offline target suggests.

\emph{IV) Misalignment.} The networks disagree on update direction. The offline target is not at all supported by the current online estimate.

We hypothesize that, all else equal, updates with higher online support are more beneficial for learning. Specifically, we posit that learning is most effective in scenarios I and II, where the recent online target fully supports the stable offline target. Conversely, scenarios III and IV introduce varying degrees of risk.  In the following, we first motivate this hypothesis empirically and then define a metric to formalize the established preference for updates with aligned targets.

\subsection{Empirical motivation}\label{sec:empirical_motivation}

\begin{figure}[t]
    \centering
    \includegraphics[width=1\linewidth]{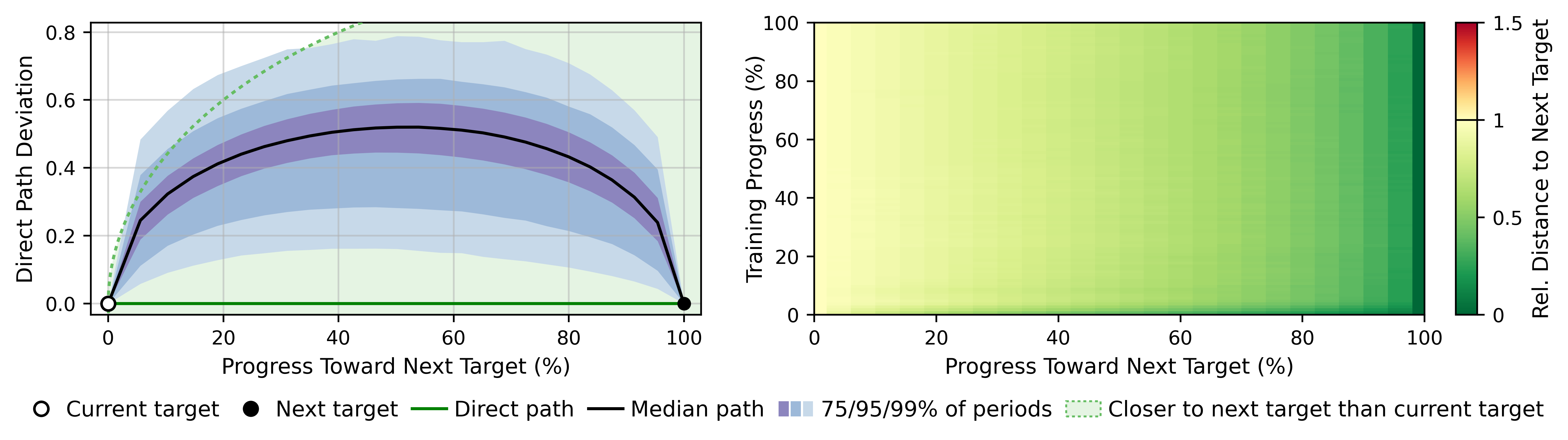}
    \vspace{-2em}
    \caption{
    How the online network moves between target-network updates (DQN over four MinAtar games, in parameter-space). \emph{Left:} Median path of the online network between target updates. Shaded bands contain the central 75\%, 95\% and 99\% of deviations. In the green region, the online network is closer to the next target than the current target is. \emph{Right:} the online network's distance to the next target, relative to the current target's distance. The online net is a better predictor of the next target than the current target in 97.8\% (SD=2.1\%; excluding endpoints) of snapshots, and improves upon the current target by 32.4\% (SD=5.2\%) on average. Further details are presented in Appendix~\ref{sec:target_to_target}.
    }
    \vspace{-1em}
    \label{fig:t2t}
\end{figure}

We subsequently elaborate why the online network's view matters when the update itself is computed from the offline target. Figure~\ref{fig:t2t} tracks the online network between two synchronizations of \gls{acr:dqn} on four \textsc{MinAtar} games. In nearly all snapshots (97.8\%, SD=2.1\%), the online network is closer to the upcoming target network than the current target network is, and on average its distance is about a third smaller. This finding suggests that the online target is usually a better indication of the next offline target than the current offline target itself.

This observation gives the four scenarios above a concrete interpretation in terms of which updates are likely to last. An update in Scenarios I or II moves the estimate toward the online target, and therefore likely also toward the next offline target, so it tends to persist. An update in Scenario IV moves the estimate in the opposite direction and is likely to be undone once the target network is synchronized. Scenario III lies in between, as only the part of the step up to the online target is likely to persist. 

Note that this reasoning is a heuristic rather than a theoretical guarantee. The online estimate is an imperfect preview, and our motivation considers each transition in isolation, whereas in deep \gls{acr:rl} an update on one transition also changes the estimates of others.

\subsection{Target alignment score}\label{ssec:alignment_score}

In the following, we formalize the intuition derived in Sections~\ref{sec:intuition} and \ref{sec:empirical_motivation} into a computable \emph{Target Alignment Score} $\mathcal{A}_t$. This score quantifies, on a per-transition basis, the degree to which the update step proposed by the stable target network is supported by the most recent online estimate. As such, it provides a principled signal for identifying updates that are both stable and up-to-date.

We first define the residual online error $\delta_t^{\text{next}}$, which approximates the online error that would remain after applying the update proposed by the offline target under unit step-size, $\delta_t^\text{next} = \delta_{t} - \bar{\delta}_t$. Using this residual, we define the base alignment score $\mathcal{A}^{\text{base}}_t \in [0, 1]$, which measures the extent to which the offline update reduces the magnitude of the current online error,

\vspace{-1em}
\begin{equation}
\mathcal{A}^{\text{base}}_t = \frac{|{\delta_t}|}{|\delta_t| + |\delta_t^\text{next}| + \varepsilon}.
\end{equation}
\vspace{-1em}

The denominator $|\delta_t| + |\delta_t^\text{next}|$ represents the total variation in error space, and $\varepsilon = 10^{-8}$ is a small constant added for numerical stability. We choose this specific fractional formulation over simpler alternatives---such as unbounded dot products or purely directional cosine similarities---because it yields a bounded $[0, 1]$ score that rigorously penalizes magnitude disagreements independent of the absolute scale of the TD errors. If the residual error $|\delta_t^\text{next}|$ is small relative to the initial error $|\delta_t|$, base alignment is high. In the ideal scenario where the offline update fully resolves the online error (i.e., $\delta_t^\text{next} \to 0$), the score is maximized (i.e., $\mathcal{A}^{\text{base}} \to 1$). Conversely, in cases of Misalignment (Scenario IV) or severe Offline Overshoot (Scenario III), the proposed update exacerbates the residual error, causing the denominator to grow relative to the numerator and diminishing the score towards zero.

However, strict adherence to this base metric would penalize offline over- and undershoots equally. Yet, as established in Section~\ref{sec:intuition}, in the case of an Offline Undershoot (Scenario II), the designated target is fully supported by the most recent estimate, while it is only partially supported in the Offline Overshoot case (Scenario III). As such, offline undershoots are inherently safer, whereas offline overshoots pose a risk of inducing parameter oscillation. To account for this asymmetry, we define the final alignment metric $\mathcal{A}_t$,

\vspace{-1em}
\begin{equation}\label{form:alignment}
\mathcal{A}_t =
\begin{cases}
    1 & \text{if } (\delta_{t} \cdot \Bar{\delta}_{t} > 0) \text{ and } (|\delta_{t}| \geq |\Bar{\delta}_{t}|) \\
    \mathcal{A}^{\text{base}}_t & \text{else}.
\end{cases}
\end{equation}
\vspace{-1em}

This metric formally captures our previous visual intuition: The conditional check isolates updates where the direction is identical and the offline magnitude is conservative. Consequently, the score equals one for Perfect Alignment (Scenario I) and Offline Undershoots (Scenario II), while falling back to the strictly penalizing $\mathcal{A}^{\text{base}}_t$ for Offline Overshoots (Scenario III) and Misalignments (Scenario IV).

\emph{Note on online target stability.} The proposed methodology relies on the alignment between online and offline targets. In the broader literature, online targets are often considered too unstable to yield valuable information. However, this instability arises primarily when the online network is used to provide bootstrapped learning targets, as this creates immediate feedback loops. In \gls{acr:tarl}, the online network is used only for directional validation. As long as the online network is updated based on stable targets provided by the offline network, its estimates remain sufficiently stable to serve as a directional reference for alignment.

\subsection{Alignment-based oversampling} \label{ssec:oversampling}

Building on the intuition that well-aligned targets yield more reliable learning signals, we introduce a simple extension to standard \gls{acr:rl} pipelines that explicitly incorporates target alignment to improve sample efficiency. We propose \emph{alignment-based oversampling}, which serves as the backbone for \gls{acr:tarl}: Given a target batch size $m$ and an oversampling margin $b \in \mathbb{N}$, we sample $m + b$ transitions from the replay buffer, compute their target alignment $\mathcal{A}_t$, and select the $m$ transitions with the highest alignment for the network update. This mechanism effectively acts as a dynamic noise filter. By temporarily omitting poorly aligned transitions, it defers their updates until subsequent network parameter changes naturally clarify the alignment signal. 

Crucially, this building block is a drop-in refinement that effortlessly converts any off-policy \gls{acr:rl} algorithm employing target networks into its target-aligned version. Algorithm~\ref{alg:tadqn} presents an instantiation for \gls{acr:tadqn}. Note the boxed lines, which highlight that \gls{acr:tarl} requires only a localized, minimal modification to the standard training loop. Alternative approaches to integrating alignment, such as prioritized sampling or loss weighting, are discussed in Appendix~\ref{sec:alternative_approaches}.

\textbf{Implicit stratification.} Unlike traditional prioritization schemes, alignment-based selection limits the persistence of its own sampling bias through regular resets: Under a hard update regime, synchronizing the target network instantly resets the target alignment to 1 across the entire replay buffer. This periodic global reset frequently reintroduces temporarily neglected transitions into the active pool with maximum priority, preventing localized biases from accumulating across target periods. 

\begin{algorithm}[h]
\caption{Target-Aligned Deep Q-Network (TA-DQN)}
\label{alg:tadqn}
\begin{algorithmic}
\STATE {\bfseries Input:} Replay buffer $\mathcal{D}$, discount factor $\gamma$, batch size $m$, oversampling margin $b$, target update interval $K$, budget $T$

\STATE Initialize online parameters $\theta$, target parameters $\Bar{\theta} \gets \theta$, and replay buffer $\mathcal{D} \gets \emptyset$

\STATE Sample initial state $S_1 \sim p$
\FOR{$t = 1, \dots, T$}
    \STATE Select action $A_t$ using $\epsilon$-greedy policy w.r.t. $Q(S_t,\cdot,\theta)$
    \STATE Execute $A_t$, observe reward $R_t$ and next state $S_{t+1}$
    \STATE Store transition $C_t = (S_t, A_t, R_t, S_{t+1}, d_t)$ in $\mathcal{D}$
    \begin{algobox}
    \STATE Sample random oversized minibatch $\{M_j\}_{j=1}^{m+b}$ from $\mathcal{D}$
    \STATE Compute offline target $\Bar{Q}_{\text{target}}(S_j)$, online target $Q_{\text{target}}(S_j)$ and alignment metric $\mathcal{A}_j$ for $M_j$
    \STATE Obtain final minibatch $\mathcal{C} \gets \text{TopK}_{m}(M, \mathcal{A})$
    \end{algobox}
    
    \STATE Update $\theta$ by minimizing loss, $L(\theta) = \frac{1}{m} \sum_{j \in \mathcal{C}} \left(\Bar{Q}_{\text{target}}(S_j) - Q(S_j, A_j, \theta) \right)^2$
    \IF{$t \pmod K = 0$}
        \STATE Update target network: $\Bar{\theta} \gets \theta$
    \ENDIF
\ENDFOR

\end{algorithmic}
\end{algorithm}

\subsection{Algorithmic intuition}
\label{sec:bias-variance}

Our hypothesis that---all else equal---updates with higher online support are more beneficial for learning can be understood through three complementary observations, each pointing to the same underlying principle: disagreement between online and offline estimates signals temporally inconsistent targets.

\emph{Avoidance of outdated targets.} Misalignment or offline overshoots suggest that the proposed offline update is no longer fully supported by the online estimate. By performing updates on those targets, we effectively chase targets which are no longer aligned with the most recent information. Avoiding such updates prevents the agent from committing to update steps that have been identified as directionally incorrect or excessively large by the most recent value estimates.

\emph{Temporal ensembling.} The online and offline networks provide complementary estimates of the same value function. While they are correlated, they function as a temporal ensemble. Lower agreement between them may indicate lower temporal consistency. Prioritizing alignment then acts as an inconsistency reduction mechanism.

\emph{Reversion of misaligned changes.} Since the online network anticipates the next target network (Section~\ref{sec:empirical_motivation}), updates that conflict with the online view are likely to be reverted after the next target network update. As such, Q-value updates based on misaligned targets induce oscillatory behavior and are potentially wasteful.

\textbf{A bias-variance perspective.} Our previous observations allow a unified interpretation through the bias-variance lens. \gls{acr:tarl}'s empirical success admits a compact reading: \emph{a high per-transition alignment score suggests that the staleness bias of the offline target has not yet manifested on a given transition}. Existing target-network schemes navigate this axis \emph{globally} via $K$ or $\tau$, whereas \gls{acr:tarl} navigates it \emph{per transition}.

We treat each bootstrapped TD residual as a noisy estimate of the expected residual $\delta_t^\star = \mathbb{E}[\delta_t \mid S_t, A_t]$, where the expectation is taken over transition noise and training stochasticity; variance is taken over the same training stochasticity, and bias is measured relative to $\delta_t^\star$, not to $Q^\star$. The offline residual $\bar{\delta}_t$ then has \emph{low variance}---$\bar{\theta}$ is held constant or smoothed over many updates---but carries \emph{bias} growing with its staleness. The online residual $\delta_t$ is the converse: continually refreshed, hence low-bias but high-variance. Traditional algorithms must select a static point along this Pareto frontier.

\gls{acr:tarl} leaves the pointwise estimator unchanged: every selected transition is regressed onto the offline target $\bar{Q}_{\text{target}}$ of the base algorithm, so the target, together with its bias and variance, is identical to that of the base algorithm. What \gls{acr:tarl} changes is which transitions enter the update. Because the selection event---agreement between $\delta_t$ and $\bar{\delta}_t$---depends on the sampled transition itself, the resulting minibatch update is outcome-dependent: its expectation is taken under an alignment-tilted rather than the uniform replay distribution, making it a biased estimate of the base algorithm's expected update. \gls{acr:tarl} deliberately accepts this selection bias in exchange for reduced staleness bias, by conditioning updates on an event correlated with low staleness bias. Decomposing the disagreement,

\vspace{-1em}
\begin{equation}
\delta_t - \bar{\delta}_t \;=\; \underbrace{(\beta_{\delta_t} - \beta_{\bar{\delta}_t})}_{\text{systematic gap}} \;+\; \underbrace{(\nu_{\delta_t} - \nu_{\bar{\delta}_t})}_{\text{stochastic gap}},
\end{equation}
\vspace{-1em}

with $\beta,\nu$ denoting bias and noise components, the small variance of $\bar{\delta}_t$ implies that small disagreement is, up to online noise, evidence that offline bias has not yet diverged on that transition. The same decomposition recovers the asymmetry of $\mathcal{A}_t$ in Eq.~\eqref{form:alignment}: Scenario~II is the low-bias regime where we accept variance reduction by retaining the conservative offline magnitude; Scenario~III flags an offline magnitude likely biased upward by staleness; Scenario~IV indicates that the directional component of the bias dominates the stochastic gap entirely.

A caveat is in order. Since $\bar{\theta}$ is a delayed copy of $\theta$, both estimators share a common ancestor and could, in principle, agree by failing in the same direction. However, agreement nonetheless remains a highly informative filter. Since the last target synchronization, the online network has absorbed gradient information from numerous transitions that the offline copy has not seen. While co-failure remains possible in deep non-convex optimization, empirical performance suggests that filtering out \emph{known} disagreements successfully prevents the agent from committing to actively conflicting, oscillatory updates. In practice, target alignment serves as a necessary, if not strictly sufficient, condition for reliable learning steps.

\section{Numerical results}\label{sec:experiments}

We evaluate \gls{acr:tarl} across both continuous and discrete control domains to assess its generality. Our experimental design is chosen to test the hypothesis that target alignment is a fundamental principle, agnostic to the specific algorithm, target update mechanism, or experience replay mechanism. We test this hypothesis in discrete control environments using \gls{acr:dqn} with uniform sampling, \gls{acr:per} and \gls{acr:reaper}, as well as in continuous control environments using \gls{acr:sac}. This selection allows us to test performance across substantially different learning dynamics, action modalities, target value formulations, experience replay strategies and target network update regimes. Specifically, we evaluate \gls{acr:tarl} in three distinct experimental settings:

\emph{Discrete control:} We evaluate the impact of target alignment in a \gls{acr:ddqn} using hard target updates on the \textsc{Atari-10} benchmark, which approximates the median score across the seminal \textsc{Atari-57} benchmark with $R^2 \approx 0.99$ \citep{aitchison_atari-5_2022}.
    
\emph{Continuous control:} We further evaluate the impact of target alignment in \gls{acr:sac} using soft target updates on six continuous control environments \citep{mujoco}.
    
\emph{Ablations:} To better understand the inner workings of \gls{acr:tarl}, its sensitivity to the oversampling margin and the target update interval, its compatibility with prioritized replay methods as well as the relevance of the transition selection criterion, we conduct additional experiments within four \textsc{MinAtar} environments \citep{minatar}.
 
 Crucially, we performed no hyperparameter tuning for \gls{acr:tarl}. We applied \gls{acr:tarl} as a direct drop-in refinement without re-tuning the base hyperparameters of the original algorithms. While we added an ablation on the \gls{acr:tarl}-specific oversampling hyperparameter $b$, we did not tune $b$ for the main experiments, but fixed $b = m$. This design choice aims to show the flexibility and robustness of \gls{acr:tarl}, and highlights its ease of integration into existing pipelines. 

 \textbf{Evaluation metric.} To evaluate learning efficiency and asymptotic performance simultaneously, our primary evaluation metric is the \gls{acr:nauc}. This metric yields a unitless score in $[0, 1]$, where $0$ indicates performance no better than a random agent throughout training, and $1$ represents achieving peak performance immediately, rewarding both sample-efficient convergence and high asymptotic returns. A formal definition is provided in Appendix~\ref{sec:nauc}.
 
 For additional details on our experimental setting, we refer to Appendix~\ref{sec:hyperparameters}. For more in-depth results and \gls{acr:nauc} tables for all experiments, we refer to Appendices~\ref{sec:nauc_results} and \ref{sec:per-seed-curves}.

\begin{figure}[b]
    \vspace{-1em}
    \centering
    \includegraphics[width=1\linewidth]{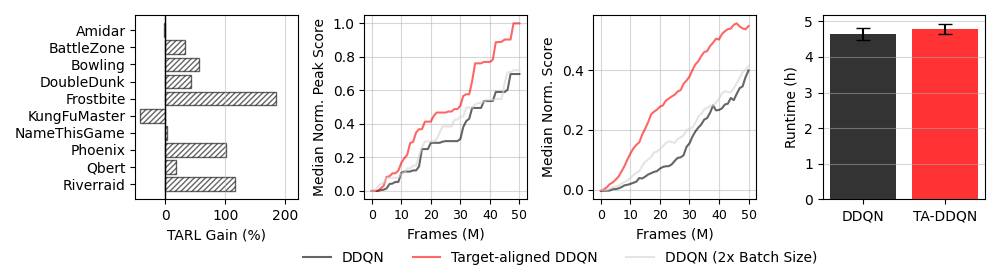}
    \vspace{-2em}
    \caption{Comparison of DDQN, Target-aligned DDQN (TA-DDQN), and DDQN with doubled batch size across the Atari-10 benchmark. (Left) Per-game peak score improvement of TA-DDQN over DDQN (\%). (Center-left) Median normalized cumulative maximum score aggregated over all games as a function of environment frames. (Center-right) Median normalized score over training, reflecting instantaneous rather than best-seen performance. (Right) Mean wall-clock runtime for DDQN and TA-DDQN. Score normalization per game is performed using the range between the random baseline and the best observed performance across all conditions. Score curves are smoothed over 10 evaluations.}
    \label{fig:atari_agg}
\end{figure}

\textbf{Discrete control.} We compared \gls{acr:ddqn} performance with and without \gls{acr:tarl} along the \textsc{Atari-10} benchmark. As displayed in Figure~\ref{fig:atari_agg}, \gls{acr:taddqn} achieves higher peak scores than its vanilla counterpart in eight out of ten games. \gls{acr:taddqn} achieved a median peak score gain of 38.18\% (95\% CI across games = [3.64, 98.69]; Mean=51.17\% (SD=63.03)). Further, \gls{acr:taddqn} scores a higher \gls{acr:nauc} on all games, raising the mean \gls{acr:nauc} by $85.1\%$ $ (0.174 \to 0.323)$. Notably, \gls{acr:taddqn} achieved this outperformance while incurring negligible increases in wall-clock time: \gls{acr:taddqn} trained for 4.78 hours (SD=0.15 hours) on average, vanilla \gls{acr:ddqn} trained for 4.63 hours (SD=0.17 hours). Thus, \gls{acr:tarl} incurred an overhead in wall-clock time of merely $3.29\%$.

To control for the confounding effect of additional network evaluations, we evaluated a vanilla \gls{acr:ddqn} using a doubled batch size. This baseline performs the same number of forward passes as \gls{acr:taddqn}, but applies backward passes to twice as many transitions. This variant performs similarly to \gls{acr:ddqn}.

\begin{figure}[t]
    \vspace{-1em}
    \centering
    \includegraphics[width=1\linewidth]{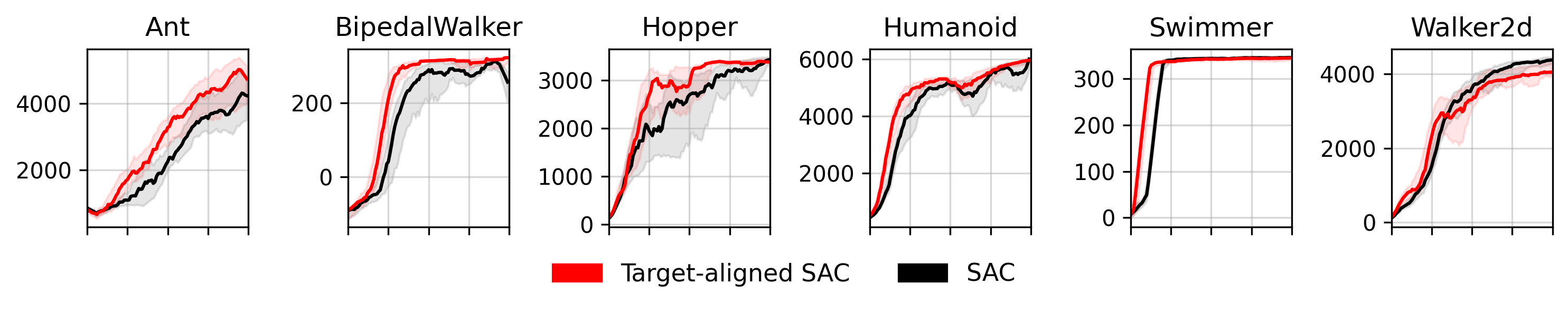}
    \vspace{-2.3em}
    \caption{Performance comparison between \gls{acr:sac} and target-aligned \gls{acr:sac} on six continuous control environments over five seeds. Curves represent median returns, smoothed over 10 evaluations, and shaded areas indicate interquartile ranges.}
    \label{fig:sac}
    \vspace{-1em}
\end{figure}

\textbf{Continuous control.}
We compared \gls{acr:sac} performance with and without \gls{acr:tarl} along six different continuous control environments, namely \textsc{Ant}, \textsc{BipedalWalker},  \textsc{Hopper}, \textsc{Humanoid}, \textsc{Swimmer}, and \textsc{Walker2d}. As in \citet{haarnoja_soft_2018}, we report performances across five seeds. As displayed in Figure~\ref{fig:sac}, \gls{acr:tasac} achieved faster convergence than \gls{acr:sac} in five out of six environments, slightly underperforming in \textsc{Walker2d} (-0.7\% \gls{acr:nauc}). Nonetheless, across all environments, TA-SAC raises the mean nAUC by $10.0\%$ $(0.640 \to 0.705)$. Unlike \gls{acr:ddqn}, where additional online forward passes can be effectively amortized via parallelization, vanilla \gls{acr:sac} does not perform online forward passes for the next states. Consequently, the naïve implementation of \gls{acr:tasac} introduces a $17.82\%$ ($SD=3.44$) increase in wall-clock training time. Yet, the performance gains in this regime highlight that even within modern algorithms under soft updates, explicit alignment filtering improves training dynamics.

\begin{figure}[b]
    \centering
    \includegraphics[width=1\linewidth]{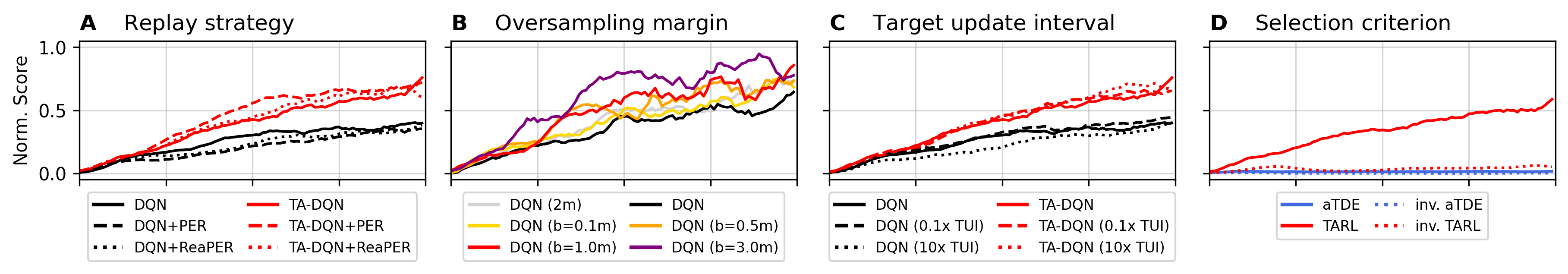}
    \vspace{-2em}
    \caption{Ablation studies of \gls{acr:tadqn} on the \textsc{MinAtar} benchmark.
    \textbf{A:} Experience replay. \gls{acr:dqn} with and without \gls{acr:tarl} under three replay strategies: uniform sampling, \gls{acr:per}, and \gls{acr:reaper}.
    \textbf{B:} Standard \gls{acr:dqn} and DQN (2m) compared against \gls{acr:tadqn} with oversampling margins $b \in \{0.1m, 0.5m, 1.0m, 3.0m\}$, denoted as DQN ($b{=}\cdot$) in the legend. Shown are the normalized score and the mean and minimum batch alignment after discarding.
    \textbf{C:} \gls{acr:dqn} and \gls{acr:tadqn} evaluated across different TUIs (default is 1{,}000).
    \textbf{D:} Selection by alignment score ($\mathcal{A}$) compared with inverted alignment ($-\mathcal{A}$), largest absolute TD error ($|\bar{\delta}_t|$), and smallest absolute TD error ($-|\bar{\delta}_t|$; $b=1m$).
    Results are aggregated across four \textsc{MinAtar} environments, with scores min-max normalized per environment before aggregation. Curves show the median over five seeds, smoothed over 10 evaluations ($m$: batch size; $b$: oversampling margin; DQN (2m): \gls{acr:dqn} with doubled batch size; TUI: target update interval; aTDE: $|\bar{\delta}_t|$).}
    \label{fig:ablations}
    \vspace{-1em}
\end{figure}

\textbf{Ablations.} All ablations were conducted across five seeds. Results are displayed in Figure~\ref{fig:ablations}. \gls{acr:nauc} results are reported in Table~\ref{tab:ablation_nauc}.

\emph{Ablation 1: Experience replay.} We compared \gls{acr:dqn} performance with and without \gls{acr:tarl} for three different experience replay strategies, namely uniform sampling, \gls{acr:per} \citep{schaul_prioritized_2015} and \gls{acr:reaper} \citep{pleiss2025reliabilityadjustedprioritizedexperiencereplay}. Target-aligned methods consistently outperform their non-aligned counterpart in every game, indicating that \gls{acr:tarl} acts as a standalone mechanism that complements, rather than substitutes, the replay strategy. Notably, \gls{acr:per} and \gls{acr:reaper} alone fall below uniform \gls{acr:dqn} ($-9.4\%$ and $-5.4\%$), but yield substantial gains over uniform \gls{acr:dqn} when combined with \gls{acr:tarl} ($+82.7\%$ and $+60.9\%$). Per-game results are displayed in Figure~\ref{fig:dqn}.

\emph{Ablation 2: Oversampling margin.} We further conducted an ablation to evaluate the impact of oversampling margin. We report the normalized scores across the environments, as well as the minimum and mean batch alignment score. Our findings suggest that performance generally increases with higher alignment: mean nAUC rises from $0.343$ ($b = 0.1m$) to $0.399$ ($b = 0.5m$), $0.420$ ($b = m$) and $0.515$ ($b = 3m$), corresponding to gains over \gls{acr:dqn} of $+17.6$, $+36.8$, $+44.0$ and $+76.4\,\%$, respectively. Naturally, minimum and mean batch alignment increase as well, and are close to saturation at around $0.9$ when $b = 3m$. Per-game results are displayed in Figure~\ref{fig:ablation_os}.

\emph{Ablation 3: Target update interval.} We evaluated the impact of different target update intervals (see Figure~\ref{fig:ablations}). We find that \gls{acr:tadqn} outperforms its non-aligned counterpart across all intervals, and that \gls{acr:tarl} helps to mitigate the adverse effects of very large update intervals: A tenfold longer target update interval lowered \gls{acr:dqn}'s mean \gls{acr:nauc} from 0.292 to 0.226 (-22.6\% relative to \gls{acr:dqn}), while \gls{acr:tadqn} was largely unaffected (0.420 $\to$ 0.411), suggesting that target alignment increases robustness to infrequent target-network updates. Per-game results are displayed in Figure~\ref{fig:tui}.

\emph{Ablation 4: Transition selection criterion.} We evaluated the impact of different transition selection criteria. We consider selection by $\mathcal{A}$ (\gls{acr:tarl}), $-\mathcal{A}$ (inverted \gls{acr:tarl}), $|\bar{\delta}_t|$ (largest absolute TD error), and $-|\bar{\delta}_t|$ (smallest absolute TD error). We find that selection by $\mathcal{A}$ is the sole criterion that facilitates stable learning ($+44.0\%$ over DQN), whereas all other criteria fall well below the DQN baseline ($-\mathcal{A}$: $-71.9\%$; $|\bar{\delta}_t|$: $-73.7\%$; $-|\bar{\delta}_t|$: $-97.0\%$). Per-game results are displayed in Figure~\ref{fig:metric_ablation}.

\section{Discussion} \label{sec:discussion}

Across a diverse set of environments, for multiple algorithms, and under both hard and soft target network update regimes, algorithms employing \gls{acr:tarl} consistently outperform their counterparts without alignment correction. Importantly, these improvements are achieved without any additional hyperparameter tuning, highlighting the robustness of the proposed approach. Moreover, \gls{acr:tarl} can be seamlessly integrated into existing \gls{acr:rl} pipelines, making it a practical drop-in enhancement that consistently improves performance across multiple domains.

While \gls{acr:tarl} improves performance, it naturally introduces computational overhead by increasing the number of forward passes required during training. However, because forward passes are relatively inexpensive and highly parallelizable in modern \gls{acr:rl} frameworks, the impact on wall-clock time remains modest given sufficient device memory, particularly in algorithms natively employing online batch forward passes like \gls{acr:ddqn} (incurring a mere 3.29\% increase for $b=m$ in Atari). Thus, to optimally balance computational costs with performance benefits, the oversampling margin $b$ should simply be scaled to device limits: Figure~\ref{fig:ablation_os} indicates that larger values of $b$ generally yield stronger gains. Consequently, we recommend setting $b$ such that the total query size $m+b$ matches the maximum single-pass batch size supported by the available hardware.

\textbf{Limitations.}
While \gls{acr:tarl} offers considerable empirical benefits, it is inherently restricted to \gls{acr:rl} algorithms that utilize target networks and does not apply to purely on-policy architectures. Further, while \gls{acr:tarl} shares structural similarities with ensemble methods, it does not yield independent estimates, nor does it constitute a measure of uncertainty. Rather, it is specifically designed to mitigate the staleness bias caused by lagged target networks (see Appendix~\ref{sec:appendix_tarl_ensembles}). Additionally, relying on oversampling requires a wider forward pass, inducing some computational overhead. Finally, like any other method that alters the sampling distribution, \gls{acr:tarl} lacks formal global guarantees against harmful i.i.d.\ violation. Providing such theoretical guarantees in deep \gls{acr:rl} currently remains a fundamental, open challenge across the entire field. However, under hard updates, the regular synchronization of the target network routinely clears the priority landscape, counteracting the buildup of systematic bias. As with other established non-uniform sampling methods like \gls{acr:per} \citep{schaul_prioritized_2015}, we rely on our extensive empirical validation, which demonstrates robust performance across diverse environments without observing any instances of degenerate state-space concentration.

\textbf{Conclusion.}
We introduced \gls{acr:tarl}, a method that prioritizes updates on transitions with high target alignment between offline and online targets. Central to our approach is a novel alignment metric that quantifies the consistency between update targets across target network updates. By leveraging this metric as a transition selection criterion, \gls{acr:tarl} promotes target recency \emph{and} stability, thereby mitigating the stability-recency trade-off induced by target networks. Empirical evaluation across a variety of benchmarks and algorithms validates \gls{acr:tarl}'s effectiveness.

\gls{acr:tarl} can be seamlessly integrated into existing \gls{acr:rl} pipelines without requiring additional hyperparameter tuning. Beyond its immediate practical benefits, our work highlights a fundamental insight: the online network---despite being too unstable to provide reliable bootstrap targets---holds highly valuable information for directional update validation. We hope \gls{acr:tarl} inspires further research into overcoming the structural bottlenecks of value-based \gls{acr:rl}. Promising future directions include the design of fully differentiable alignment-aware loss functions and the exploration of alignment signals for dynamic representation learning.

\clearpage

\begin{ack}
L.S.P. was supported by a Roland Berger doctoral fellowship. Roland Berger had no role in the design, execution, analysis, or reporting of this research. The authors declare no competing interests.
\end{ack}

\bibliographystyle{plainnat}
\bibliography{references}


\newpage
\appendix

\section{Alternative integration strategies}\label{sec:alternative_approaches}

Several alternatives to oversampling are worth considering to integrate alignment into the learning process:

\emph{Threshold-based filtering.}
A natural alternative to our top-$k$ oversampling scheme is to discard all transitions whose alignment score falls below a fixed threshold and to update on whatever remains. We deliberately opted against this design. A threshold would reintroduce exactly the kind of hyperparameter that \gls{acr:tarl} is designed to avoid. Further, it would introduce variable batch sizes, which would confound our comparisons: by keeping the number of backward passes identical to the baselines, we ensure that observed gains are attributable to \emph{which} transitions are selected rather than to \emph{how many} updates are performed. Variable batch size would further cause a practical challenge, as it would lead to inconsistent accelerator utilization and memory allocation.

\emph{Prioritized sampling.} Replay transitions could be sampled based on alignment, favoring well-aligned updates. However, this would require computing alignment for the full buffer, which is computationally prohibitive. Approximate solutions, such as maintaining stored alignment values updated on sampling (analogous to Prioritized Experience Replay \citep{schaul_prioritized_2015}), are likely not applicable, as the alignment metric relies on frequently changing online estimates.

\emph{Loss weighting.} Alignment can be used as a weight within the loss function, upweighting well-aligned samples and downweighting poorly aligned ones. While flexible, this approach requires hyperparameter tuning and may interfere with other loss-weighting schemes (e.g., importance sampling in PER), thereby limiting the method's general applicability.

\emph{Loss scaling.} Alternatively, alignment may be used to directly scale the loss, effectively modulating the learning rate on a per-sample basis. This is conceptually clean but risks destabilization unless normalization or re-scaling strategies are applied, requiring careful hyperparameter tuning.

We consider oversampling the most robust and conceptually sound approach, as it does not require hyperparameter tuning, is easily integrated into existing pipelines and ensures target alignment while causing limited computational overhead. We therefore deliberately chose to limit this paper's scope to oversampling. Yet, we plan to pursue other directions in future work.

\section{Distinctions from ensemble methods}
\label{sec:appendix_tarl_ensembles}

\gls{acr:tarl} shares superficial structural similarities with ensemble methods, as both evaluate multiple network estimates. However, the underlying mechanisms, computational costs, and theoretical drivers of \gls{acr:tarl} are fundamentally distinct. This section clarifies why \gls{acr:tarl}'s performance gains may not be explained by generic ensemble-based variance reduction and are instead explained by the mitigation of the stability-recency trade-off.

Standard ensemble methods rely on independently initialized and/or trained networks to achieve variance reduction. \gls{acr:tarl} diverges from this paradigm in several critical ways:

\emph{Lack of independent estimators:} The target network in \gls{acr:tarl} is not an independent estimator; it is strictly a lagged clone of the online network. They are identically initialized, and the sole mathematical and informational difference between them is the parameter updates derived since the last synchronization---a delta we define as the \textit{recency gap}.

\emph{Asymmetric metric design:} \gls{acr:tarl} does not strictly prioritize for consensus, but instead, \gls{acr:tarl} employs an asymmetric alignment score to specifically target directional temporal consistency.

\emph{Computational efficiency:} Ensembles demand considerable overhead, as they require maintaining multiple networks and additional backward passes. \gls{acr:tarl} exploits existing architectures, requiring only wider forward passes. This results in a significantly reduced computational footprint (less than a 4\% increase in wall-clock time on Atari).

Further, because the recency gap is the exclusive structural difference between the online network and the target network, it is the only variable that allows the online network to meaningfully alter the target network's predictions. The gains observed in \gls{acr:tarl} are therefore explained by recency, not by variance reduction.

The functional distinction between \gls{acr:tarl} and generic ensembling is further supported by two empirical observations. First, \gls{acr:sac} natively employs a twin-critic ensemble to mitigate overestimation bias. If the benefit of \gls{acr:tarl} stemmed merely from combining multiple value estimates, it would be largely redundant in this setup. Instead, integrating \gls{acr:tarl} into twin-critic \gls{acr:sac} raises the mean \gls{acr:nauc} by $10.0\%$ and improves performance in five of six environments, indicating that mitigating the staleness bias operates orthogonally to traditional ensembling. Second, the target update interval ablation (Table~\ref{tab:ablation_nauc}) directly manipulates the recency gap. A tenfold longer interval lowers the mean \gls{acr:nauc} of \gls{acr:dqn} by $22.6\%$, whereas \gls{acr:tadqn} loses only $2.1\%$. Consequently, the advantage of \gls{acr:tarl} over \gls{acr:dqn} grows with the interval, from $+40.1\%$ ($0.1\times$) to $+44.0\%$ ($1\times$) and $+82.2\%$ ($10\times$). The gains of \gls{acr:tarl} are thus largest where staleness is most severe, and \gls{acr:tarl} removes nearly all of the degradation that staleness inflicts on \gls{acr:dqn}, a pattern expected from a mechanism that targets the recency gap.

\section{Experimental design}\label{sec:hyperparameters}

Hyperparameters for \gls{acr:sac} were set to the standards defined in the RL Baselines3 Zoo \citep{rl-zoo3}. In \textsc{Ant},  \textsc{Hopper}, \textsc{Swimmer}, and \textsc{Walker2d}, we trained for 1 million timesteps per run. In \textsc{BipedalWalker}, we trained for 500{,}000 steps. In \textsc{Humanoid}, we trained for 2 million timesteps per run. \gls{acr:dqn} hyperparameters for \textsc{MinAtar} were set to recommendations published within \citet{minatar_params}. We trained for 2 million timesteps per run. \gls{acr:ddqn} hyperparameters for \textsc{Atari} were set in accordance with seminal work \citep{mnih_human-level_2015, schaul_prioritized_2015, vanhasselt2015deepreinforcementlearningdouble}. We trained for 50 million frames (12.5 million timesteps; one seed per game). \gls{acr:per} and \gls{acr:reaper} were configured as in the original work \citep{schaul_prioritized_2015, pleiss2025reliabilityadjustedprioritizedexperiencereplay}. Importance sampling weights are batch-normalized by division with the maximum weight within the (filtered) batch. When using \gls{acr:tarl} with \gls{acr:per} or \gls{acr:reaper}, sampling priorities are updated for the enlarged batch. Oversampling margins are, when required, converted to integers via flooring (e.g., when $m=32$ and $b=0.1m$, $b \gets 3$). Atari experiments employ the standard modification steps extensively described in seminal work \citep{mnih_human-level_2015, vanhasselt2015deepreinforcementlearningdouble, schaul_prioritized_2015}, including frame skips, grayscaling and resizing, reward clipping, no-op starts and terminal-on-life-loss. For \gls{acr:dqn} and \gls{acr:ddqn}, we employ the standard network architecture described in \citet{mnih_human-level_2015}. \gls{acr:sac} employs ReLU activations and the \textsc{StableBaselines3} \citep{antonin_raffin_stable-baselines3_2024} standard architecture: two hidden layers of 256 units each (except BipedalWalker, which employs hidden layers of 400 and 300 units).

\begin{table*}[h]
\vspace{-1em}
  \centering
  
  \begin{minipage}[t]{0.32\linewidth}
    \caption{Hyperparameters for \gls{acr:sac} on continuous control env.\protect\footnotemark}
    \label{tab:models}
    \centering
    \resizebox{\linewidth}{!}{%
      \begin{sc}
        \begin{tabular}{lc}
          \toprule
            Parameter &  Value \\
          \midrule
            Learning rate & $3e-4$ \\
            Buffer size & $1e6$ \\
            Batch size & $256$ \\
            Learning starts & $1e4$ \\
            Discount factor & $0.99$ \\
            Soft update coeff. & $0.005$ \\
            Update per steps & $1$ \\
            \# Gradient steps & $1$ \\
            Optimizer & Adam \\
          \bottomrule
        \end{tabular}
      \end{sc}
    }
  \end{minipage}%
  \hfill
  \begin{minipage}[t]{0.32\linewidth}
    \caption{Hyperparameters for \gls{acr:dqn} on MinAtar.}
    \label{tab:models2}
    \centering
    \resizebox{\linewidth}{!}{%
      \begin{sc}
        \begin{tabular}{lc}
          \toprule
            Parameter & Value \\
          \midrule
            Learning rate & $2.5e-4$ \\
            Buffer size & $1e5$ \\
            Batch size & $32$ \\
            Learning starts & $5{,}000$ \\
            Discount factor & $0.99$ \\
            Target update int. & $1{,}000$ \\
            Starting expl. & $1.0$ \\
            Expl. fraction & $0.05$ \\
            Final expl. & $0.01$ \\
            Optimizer & RMSprop \\
          \bottomrule
        \end{tabular}
      \end{sc}
    }
  \end{minipage}%
  \hfill
  \begin{minipage}[t]{0.32\linewidth}
    \caption{Hyperparameters for \gls{acr:ddqn} on Atari.}
    \label{tab:ddqn_hyperparams}
    \centering
    \resizebox{\linewidth}{!}{%
      \begin{sc}
        \begin{tabular}{lc}
          \toprule
            Parameter & Value \\
          \midrule
            Learning rate & $2.5e-4$ \\
            Total frames & $50{,}000{,}000$ \\
            Buffer size & $1e6$ \\
            Batch size & $32$ \\
            Learning starts & $50{,}000$ \\
            Discount factor & $0.99$ \\
            Target update int. & $30{,}000$ \\
            Starting expl. & $1.0$ \\
            Expl. fraction & $0.08$ \\
            Final expl. & $0.01$ \\
            Eval expl. fraction & $0.001$ \\
            Train frequency & $4$ \\
            Gradient steps & $1$ \\
            Frame stack & $4$ \\
            Optimizer & RMSprop \\
          \bottomrule
        \end{tabular}
      \end{sc}
    }
  \end{minipage}
  \vskip 0.1in
\end{table*}
\vspace{-2em}
\footnotetext{In Swimmer-v4, a discount factor of $0.999$ was used. For BipedalWalker-v3, we employ generalized state-dependent exploration, a learning rate of 7.3e-4, a buffer size of 3e5, a discount factor of 0.98, a soft update coefficient of 0.02, and perform 64 gradient steps every 64 environment steps.}

\paragraph{Evaluation.} For all environments, $100$ evaluations were evenly spaced throughout the training procedure. Each agent evaluation consisted of five full trajectories in the environment, going from initial to terminal state. The evaluation score of a single agent evaluation is the average total score across those five evaluation trajectories. \gls{acr:sac} employs deterministic action selection during evaluation.

\section{Implementation details}

\paragraph{TA-DDQN} The implementation of \gls{acr:taddqn} is identical to \gls{acr:tadqn}, except that it uses the established \gls{acr:ddqn} offline target, $\Bar{Q}_{\text{target}}(S_t) = R_{t} + (1 - d_{t}) \cdot \gamma \cdot Q\left(S_{t+1}, \arg\max_{a'} Q(S_{t+1}, a'; \theta); \Bar{\theta}\right)$ instead of the \gls{acr:dqn} offline target for alignment calculation and parameter updates.

\paragraph{TA-SAC} To adapt the alignment metric for the \gls{acr:sac} algorithm, we make specific adjustments to account for both the twin-critic architecture and the maximum entropy objective. For the twin-critic setup, the \gls{acr:td} errors and subsequent alignment scores are first computed independently for each of the two Q-networks. To ensure a conservative evaluation of a transition's utility and prevent overestimation, the final alignment score assigned to each transition is the minimum of the alignment scores across the twin critics. The stochastic policy's entropy term is integrated into the metric by incorporating it directly into the \gls{acr:td} targets. Specifically, the scaled log-probability of the next action, $\alpha \log \pi(a'|s')$, is subtracted from the next-state Q-value estimates for both the online and offline target calculations. This ensures that the alignment calculation evaluates the critics based on the standard soft Bellman equations, requiring no fundamental structural changes to the alignment formulation itself.

\section{Normalized area under the curve}\label{sec:nauc}

To evaluate learning efficiency and asymptotic performance simultaneously, we report the \gls{acr:nauc}. For a given environment, let $R_t$ denote the mean evaluation return at step $t$, and let $R_{\text{min}}$ and $R_{\text{max}}$ represent the environment-specific random baseline and the maximum return achieved across all compared algorithms, respectively. To ensure the metric focuses on learning progress relative to the baseline and remains robust to ``cold start'' noise, we calculate the \gls{acr:nauc} by clipping normalized scores at zero and averaging over the total number of evaluation checkpoints $T$,

\vspace{-1em}
\begin{equation}
\text{nAUC} = \frac{1}{T} \sum_{t=1}^{T} \max \left( 0, \frac{R_t - R_{\text{min}}}{R_{\text{max}} - R_{\text{min}}} \right)
\end{equation}
\vspace{-1em}

This metric provides a within-study relative, unitless score in the range $[0, 1]$, where a score of $0$ indicates performance no better than a random agent throughout training, and a score of $1$ represents achieving the peak performance observed across all conditions at the very first evaluation checkpoint, and maintaining it consistently ever after. We report \gls{acr:nauc} across all experimental conditions as a comparable metric of sample efficiency.

\section{Experiment on target-to-target trajectories}\label{sec:target_to_target}

Figure~\ref{fig:t2t} displays how \gls{acr:dqn}'s online network moves during a target period, the interval between two consecutive target-network updates. We train \gls{acr:dqn} on the four \textsc{MinAtar} games \textsc{Asterix}, \textsc{Breakout}, \textsc{Freeway}, and \textsc{Space Invaders}, according to the hyperparameter settings described in Appendix~\ref{sec:hyperparameters}, with 5 seeds per game.

Within each target update interval, we record the online network every 40 environment steps. This gives 26 snapshots per period, wherein the first is the current target, the last is the next target. Each snapshot stores the full parameter vector of the online network (all weights and biases, flattened). We record $1{,}037{,}400$ individual measurements ($957{,}600$ excluding endpoints) across $39{,}900$ target periods, stratified over training time. For the heatmap, we average the relative distance over all periods within 100 equal bins of training time and over seeds, then over games with equal weight.

Each snapshot is described relative to its period's direct path, using three quantities: \emph{Progress} is the projection of the snapshot's displacement from the current target onto the direct path, divided by the length of the direct path. It is 0 at the current target and 1 at the next target. \emph{Deviation} is the snapshot's distance from the straight line through both targets, divided by the same length. \emph{Relative distance to the next target} describes the snapshot's distance to the next target, divided by the distance between the two targets. Values below 1 mean the online network is closer to the next target than the current target is. Every quantity is relative to the distance between consecutive targets. Periods from different games, stages of training and seeds therefore share one scale and can be pooled.

\section{Normalized area under the curve tables}\label{sec:nauc_results}

 In Tables~\ref{tab:atari_10_nauc}--\ref{tab:ablation_nauc}, gains denote the relative change in \gls{acr:nauc} over the respective baseline (DDQN, SAC, and DQN). Per-environment gains compare both methods within a single environment, whereas aggregate gains compare their mean \gls{acr:nauc} across environments, i.e., $\left(\overline{\text{nAUC}}_{\text{method}} / \overline{\text{nAUC}}_{\text{baseline}} - 1\right) \cdot 100$. Averaging per-environment relative gains instead yields considerably larger values (e.g., $+152\%$ rather than $+85\%$ on \textsc{Atari-10}, and $+74\%$ rather than $+44\%$ for \gls{acr:tadqn} on \textsc{MinAtar}). These values are dominated by environments with near-zero baseline \gls{acr:nauc}; we therefore report the relative change between mean \gls{acr:nauc} values as the more conservative estimate.
 
\begin{table}[!ht]
  \centering
  \renewcommand{\arraystretch}{0.84}   
  \setlength{\abovecaptionskip}{1pt}
  \setlength{\belowcaptionskip}{2pt}
 
  \caption{Normalized area under the curve for Atari-10.}
  \label{tab:atari_10_nauc}
  \begin{small}
    \begin{sc}
      \begin{tabular}{lccc}
        \toprule
          Environment & DDQN & TA-DDQN & Gain (\%) \\
        \midrule
          Amidar         & 0.325 & 0.414 & +27.4 \\
          BattleZone     & 0.088 & 0.270 & +205.8 \\
          Bowling        & 0.035 & 0.193 & +446.5 \\
          DoubleDunk     & 0.108 & 0.296 & +175.0 \\
          Frostbite      & 0.077 & 0.338 & +337.0 \\
          KungFuMaster   & 0.239 & 0.298 & +24.7 \\
          NameThisGame   & 0.341 & 0.439 & +28.7 \\
          Phoenix        & 0.113 & 0.255 & +126.5 \\
          Qbert          & 0.201 & 0.361 & +79.3 \\
          Riverraid      & 0.216 & 0.363 & +67.9 \\
        \midrule
          \textbf{Overall Average} & \textbf{0.174} & \textbf{0.323} & \textbf{+85.1} \\
        \bottomrule
      \end{tabular}
    \end{sc}
  \end{small}
 
  \vspace{0.9em}
 
  \caption{Normalized area under the curve for continuous control environments.}
  \label{tab:mujoco_nauc_results}
  \begin{small}
    \begin{sc}
      \begin{tabular}{lccc}
        \toprule
          Environment & SAC & TA-SAC & Gain (\%) \\
        \midrule
          Ant-v4           & 0.452 & 0.534 & +18.2 \\
          BipedalWalker-v3 & 0.657 & 0.784 & +19.2 \\
          Hopper-v4        & 0.630 & 0.715 & +13.5 \\
          Humanoid-v4      & 0.660 & 0.733 & +11.2 \\
          Swimmer-v4       & 0.857 & 0.879 & +2.5 \\
          Walker2d-v4      & 0.587 & 0.583 & -0.7 \\
        \midrule
          \textbf{Overall Average} & \textbf{0.640} & \textbf{0.705} & \textbf{+10.0} \\
        \bottomrule
      \end{tabular}
    \end{sc}
  \end{small}
 
  \vspace{0.9em}
 
  \begingroup
  \caption{Normalized area under the learning curve for all ablations in Figure~\ref{fig:ablations}.
    Evaluation returns are min-max normalized per game across all runs in this table to enable comparability within each column.}
  \label{tab:ablation_nauc}
  \small
  \setlength{\tabcolsep}{4pt}
  \begin{tabular}{@{}l ccccc r@{}}
    \toprule
    Condition & Asterix & Breakout & Freeway & SpaceInv. & Mean & Gain (\%) \\
    \midrule
    DQN              & 0.090 & 0.368 & 0.482 & 0.227 & 0.292 & -- \\
    DQN ($2m$)       & 0.096 & 0.443 & 0.530 & 0.290 & 0.340 & $+16.5$ \\
    TA-DQN           & 0.272 & 0.474 & 0.631 & 0.304 & 0.420 & $+44.0$ \\
    \midrule
    \multicolumn{7}{@{}l}{\textit{(A) Replay strategy}} \\
    DQN+PER          & 0.066 & 0.401 & 0.345 & 0.245 & 0.264 & $-9.4$ \\
    DQN+ReaPER       & 0.075 & 0.376 & 0.411 & 0.242 & 0.276 & $-5.4$ \\
    TA-DQN+PER       & 0.506 & 0.522 & 0.647 & 0.457 & 0.533 & $+82.7$ \\
    TA-DQN+ReaPER    & 0.394 & 0.480 & 0.630 & 0.374 & 0.470 & $+60.9$ \\
    \midrule
    \multicolumn{7}{@{}l}{\textit{(B) Oversampling margin}} \\
    TA-DQN ($b{=}0.1m$)   & 0.108 & 0.439 & 0.565 & 0.260 & 0.343 & $+17.6$ \\
    TA-DQN ($b{=}0.5m$)   & 0.195 & 0.451 & 0.616 & 0.335 & 0.399 & $+36.8$ \\
    TA-DQN ($b{=}3.0m$)   & 0.480 & 0.470 & 0.727 & 0.382 & 0.515 & $+76.4$ \\
    \midrule
    \multicolumn{7}{@{}l}{\textit{(C) Target update interval}} \\
    DQN (0.1$\times$ TUI)     & 0.089 & 0.394 & 0.500 & 0.238 & 0.305 & $+4.6$ \\
    DQN (10$\times$ TUI)      & 0.050 & 0.281 & 0.346 & 0.226 & 0.226 & $-22.6$ \\
    TA-DQN (0.1$\times$ TUI)  & 0.312 & 0.486 & 0.603 & 0.310 & 0.428 & $+46.6$ \\
    TA-DQN (10$\times$ TUI)   & 0.303 & 0.452 & 0.603 & 0.287 & 0.411 & $+41.0$ \\
    \midrule
    \multicolumn{7}{@{}l}{\textit{(D) Selection criterion}} \\
    aTDE             & 0.007 & 0.005 & 0.276 & 0.019 & 0.077 & $-73.7$ \\
    inv.\ aTDE       & 0.007 & 0.022 & 0.002 & 0.004 & 0.009 & $-97.0$ \\
    inv.\ TARL       & 0.013 & 0.062 & 0.198 & 0.055 & 0.082 & $-71.9$ \\
    \bottomrule
  \end{tabular}
  \par
  \endgroup
\end{table}
\clearpage
 
\section{Additional figures}\label{sec:per-seed-curves}

\begin{figure}[h!]
    \centering
    \includegraphics[width=1\linewidth]{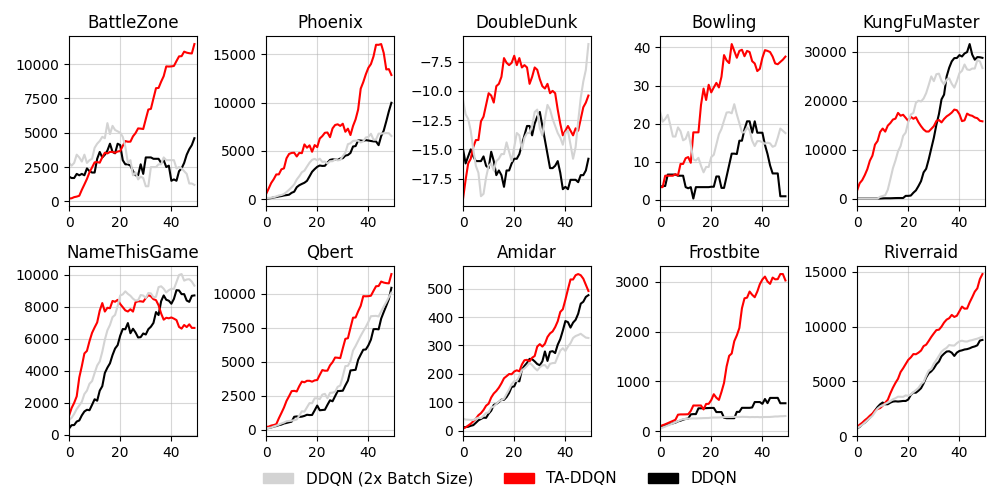}
    \vspace{-2em}
    \caption{Per-game performance comparison between \gls{acr:ddqn}, \gls{acr:ddqn} with doubled batch size and \gls{acr:taddqn} on the Atari-10 benchmark. Curves represent returns, smoothed over 10 evaluations.}
    \label{fig:enter-label}
    \vspace{-1em}
\end{figure}

\begin{figure}[h!]
    \centering
    \includegraphics[width=1\linewidth]{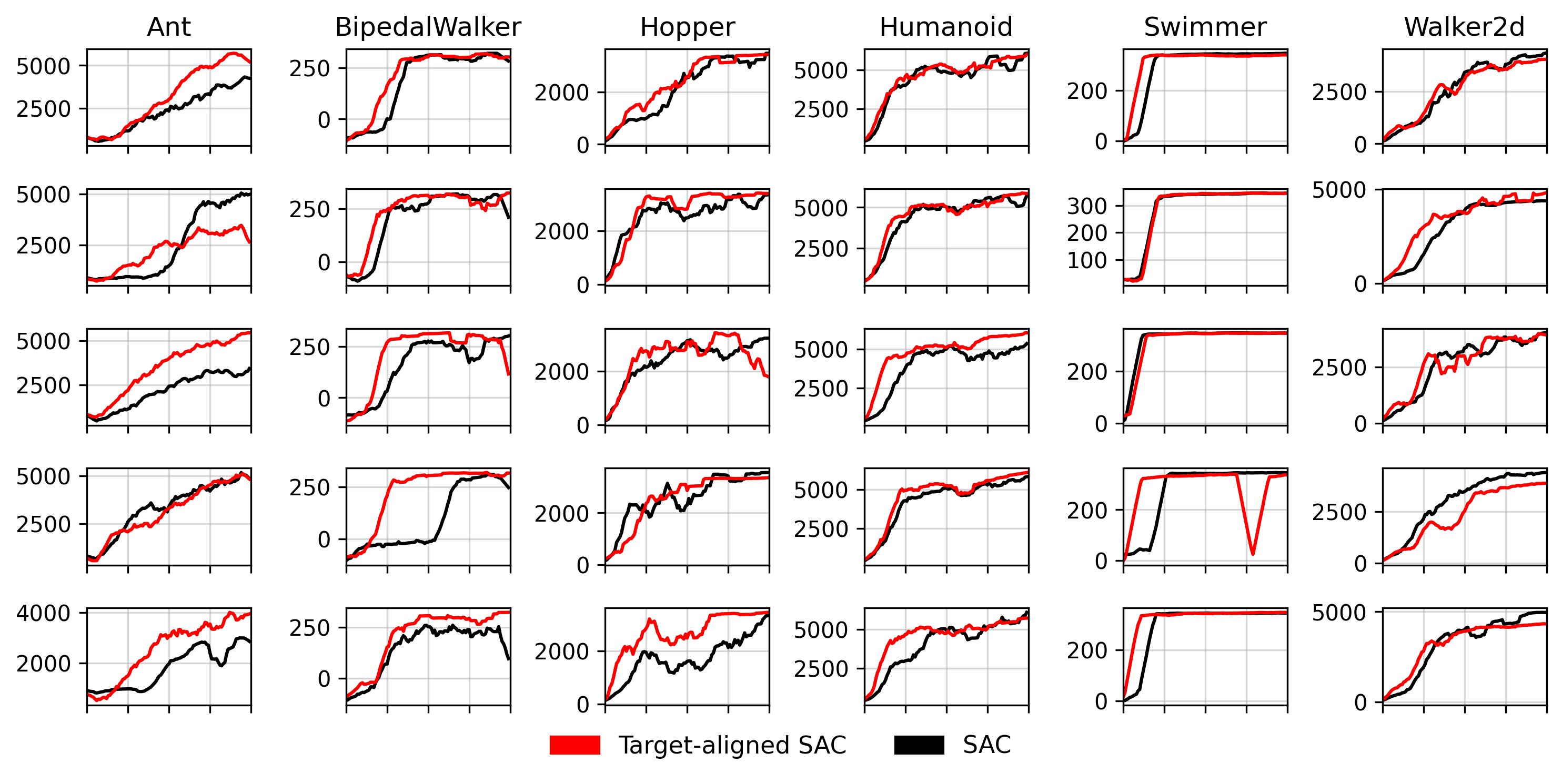}
    \vspace{-2em}
    \caption{Per-seed performance comparison between \gls{acr:sac} and \gls{acr:tasac} on six continuous control environments. Curves represent returns, smoothed over 10 evaluations.}
    \label{fig:sac_per_seed}
\end{figure}

\begin{figure}[h!]
    \centering
    \includegraphics[width=1\linewidth]{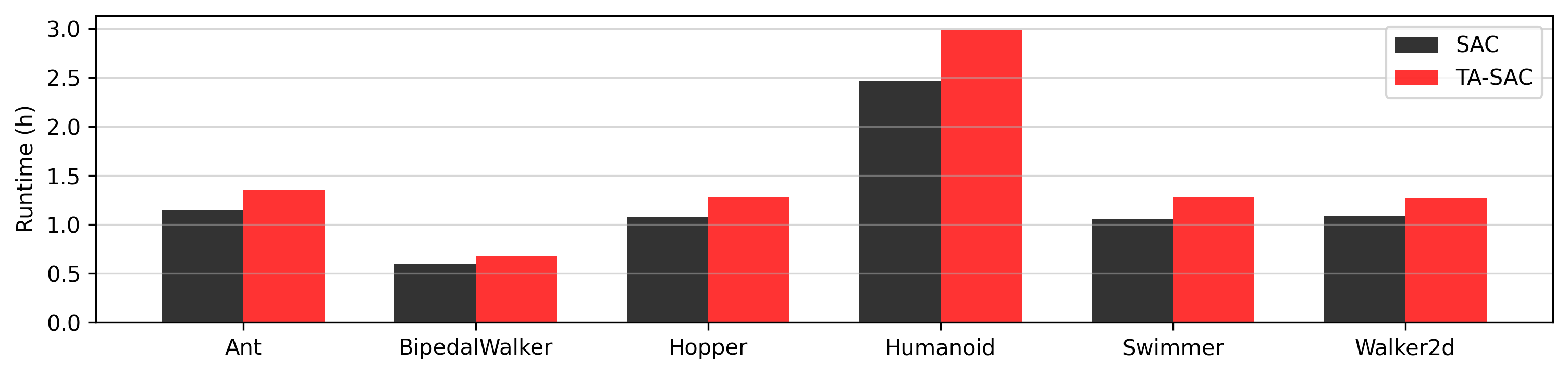}
    \vspace{-2em}
    \caption{Wall-clock training time for \gls{acr:sac} and \gls{acr:tasac} on six continuous control environments.}
    \label{fig:sac_runtime}
\end{figure}

\begin{figure}[h!]
    \centering
    \includegraphics[width=1\linewidth]{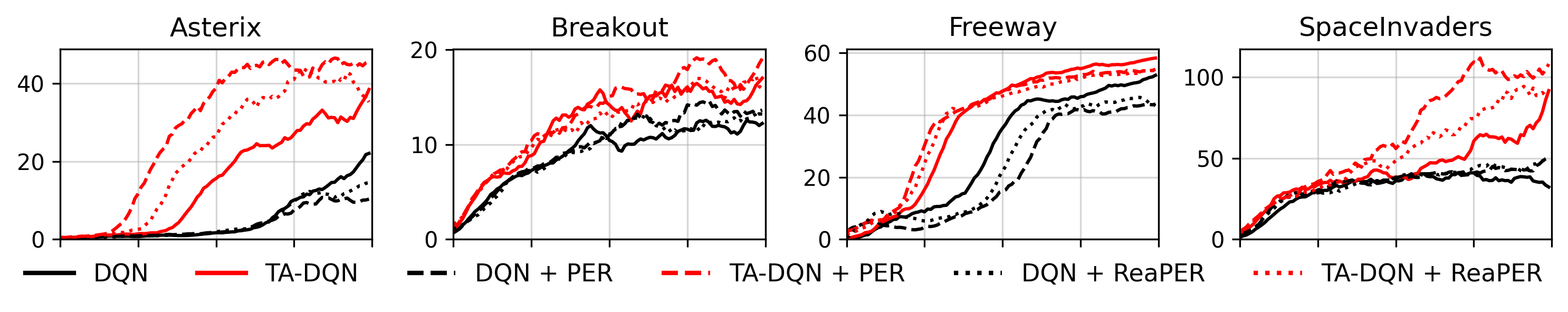}
    \vspace{-2em}
    \caption{Performance comparison between \gls{acr:dqn} and target-aligned \gls{acr:dqn} under different experience replay strategies on four games of the \textsc{MinAtar} benchmark over five seeds. Curves represent median returns, smoothed over 10 evaluations.}
    \label{fig:dqn}
\end{figure}

\begin{figure}[h!]
    \centering
    \includegraphics[width=1\linewidth]{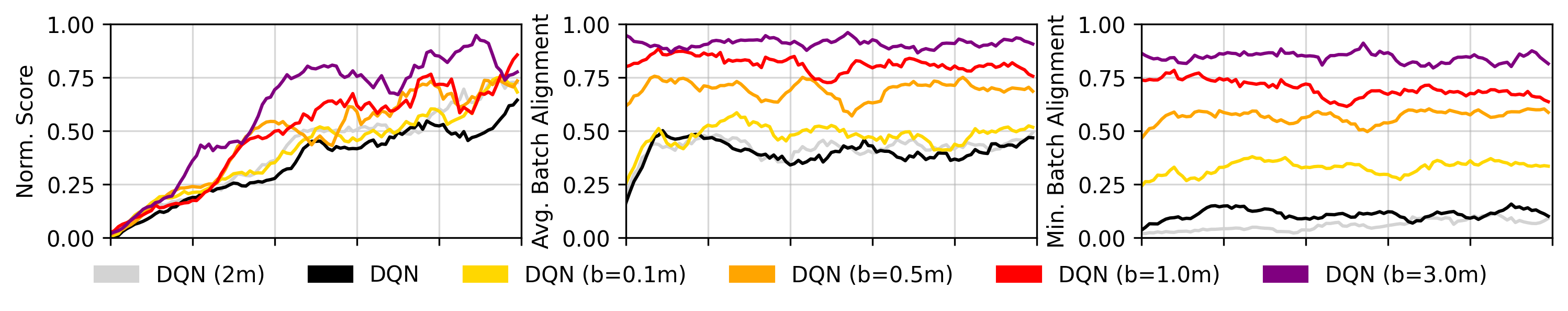}
    \vspace{-2em}
    \caption{Impact of the oversampling margin on performance and batch alignment. Results are aggregated across four MinAtar environments. The subplots compare standard DQN and a DQN with doubled batch size---denoted as DQN (2m)---against \gls{acr:tadqn}, evaluated at oversampling margins $b \in \{0.1m, 0.5m, 1.0m, 3.0m\}$, denoted as DQN ($b{=}\cdot$) in the legend. All curves are smoothed over 10 evaluations. (Left) Median normalized evaluation scores. Scores were min-max normalized per environment prior to aggregation. (Center) Average batch alignment post-discard. (Right) Minimum batch alignment post-discard.}
    \label{fig:ablation_os}
\end{figure}

\begin{figure}[h!]
    \centering
    \includegraphics[width=1\linewidth]{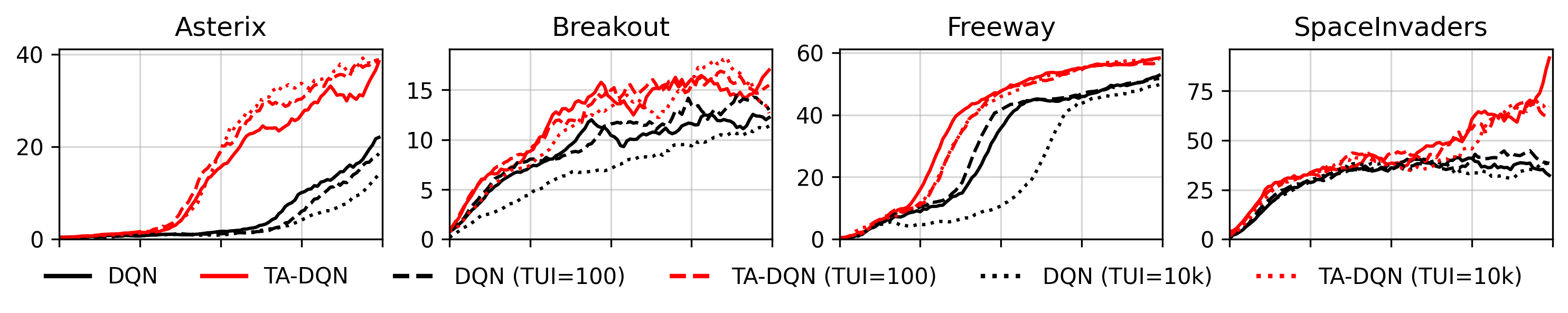}
    \vspace{-2em}
    \caption{Impact of the target update interval. All curves are smoothed over 10 evaluations (TUI = target update interval).}
    \label{fig:tui}
\end{figure}

\begin{figure}[h!]
    \centering
    \includegraphics[width=1\linewidth]{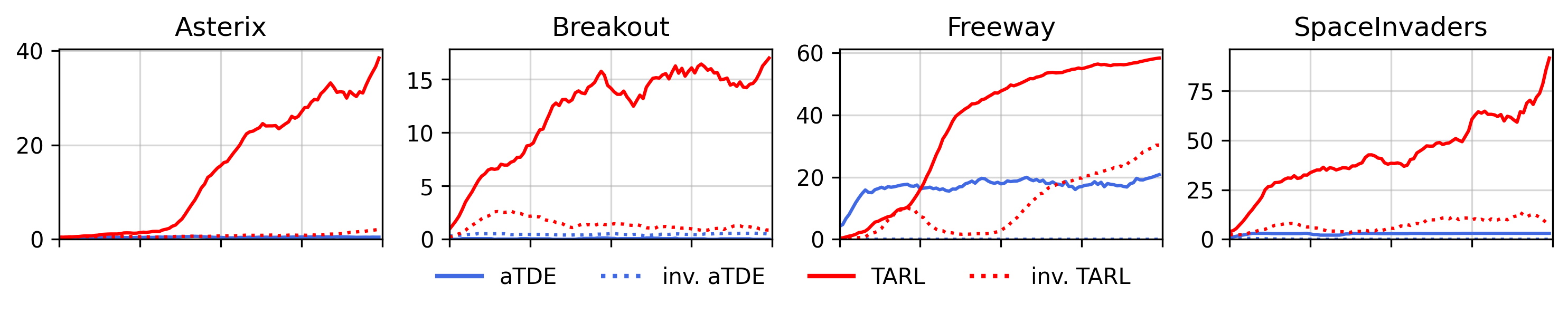}
    \vspace{-2em}
    \caption{Impact of the chosen metric. All curves are smoothed over 10 evaluations. Batches in each condition were generated by sampling 2m candidates and selecting the top m candidates. Selection metrics were aTDE ($|\bar{\delta}_t|$), inv. aTDE ($-|\bar{\delta}_t|$), TARL ($\mathcal{A}$) and inv. TARL ($-\mathcal{A}$).}
    \label{fig:metric_ablation}
\end{figure}

\clearpage

\section{Target alignment and upstream error propagation}
\label{sec:theory-error-propagation}

We illustrate the effect of \gls{acr:tarl} on error propagation in a minimal tabular MDP in which all quantities can be computed in closed form.
The example is not a general guarantee; it shows how successor-value drift enters the selection criterion and how this changes the propagation of stale target values when the target network is synchronized infrequently.

\subsection{The mechanics of the alignment metric}
\label{subsec:alignment-mechanics}

For the online and lagged targets
\begin{equation}
  y = r + \gamma \max_{a'} Q(s',a', \theta), \qquad
  \bar{y} = r + \gamma \max_{a'} Q(s',a', \Bar{\theta}),
\end{equation}
with residuals $\delta = y - Q(s,a)$ and $\bar{\delta} = \bar{y} - Q(s,a)$, the difference between the two evaluates to
\begin{equation}
  \label{eq:drift}
  \delta - \bar{\delta}
  = \gamma \Bigl[ \max_{a'} Q(s',a', \theta) - \max_{a'} Q(s',a', \Bar{\theta}) \Bigr].
\end{equation}
Because the reward and the current-state estimate cancel, the only information that \gls{acr:tarl} adds to the standard residual $\bar{\delta}$ is the drift of the successor value between the online and target critics; it does not directly measure target bias or uncertainty.
The top-$m$ mechanism modulates update frequencies based on this drift without altering the TD targets used for the selected transitions.
For terminal transitions, $y = \bar{y} = r$ and hence $\delta = \bar{\delta}$.

In the analysis below, we only use the following property of the alignment score $\mathcal{A}(\delta, \bar{\delta})$: for $\bar{\delta} \neq 0$, the score attains its maximum $\mathcal{A}(\bar{\delta}, \bar{\delta}) = 1$, and $\mathcal{A}(\delta, \bar{\delta}) < 1$ whenever $\delta \bar{\delta} \le 0$ or $|\delta| < |\bar{\delta}|$.
That is, a transition is less aligned when the online target contradicts the step prescribed by the lagged target or supports only part of it.

\subsection{Uniform sampling and the stale successor problem}
\label{subsec:stale-successor}

We analyze a minimal two-step chain with a single action:
\begin{equation}
  s_0 \xrightarrow{\;0\;} s_1 \xrightarrow{\;R\;} \bot .
\end{equation}
We write $u = Q(s_0)$ and $v = Q(s_1)$, whose true values are $\gamma R$ and $R$, and define the value errors as
\begin{equation}
  e_u = u - \gamma R, \qquad e_v = v - R.
\end{equation}

Let $\bar{u}, \bar{v}$ denote the target-network values, equal to $u, v$ at the start of each interval. Each update is a tabular step with learning rate $\eta \in (0,1)$ towards the lagged target, and the target network is synchronized every $K$ updates.
We write $e_{u,j}$ and $e_{v,j}$ for the errors at the start of the $j$-th target interval.
During a frozen-target interval, if $u$ and $v$ are updated $n_u$ and $n_v$ times, respectively, their updated values become
\begin{align}
  u' &= (1-\eta)^{n_u} u + \bigl[1 - (1-\eta)^{n_u}\bigr] \gamma \bar{v}, \\
  v' &= (1-\eta)^{n_v} v + \bigl[1 - (1-\eta)^{n_v}\bigr] R.
\end{align}
Under uniform sampling, each update selects either transition with probability $1/2$, so that $n_u, n_v \sim \mathrm{Binomial}(K, 1/2)$.
Since $\mathbb{E}\bigl[(1-\eta)^{n}\bigr] = (1-\eta/2)^K$ for $n \sim \mathrm{Binomial}(K, 1/2)$, the expected errors evolve according to the linear system
\begin{equation}
  \mathbb{E}
  \begin{bmatrix} e_{u,j+1} \\ e_{v,j+1} \end{bmatrix}
  =
  \begin{bmatrix} \lambda & \gamma(1-\lambda) \\ 0 & \lambda \end{bmatrix}
  \mathbb{E}
  \begin{bmatrix} e_{u,j} \\ e_{v,j} \end{bmatrix},
  \qquad \lambda = (1 - \eta/2)^K .
\end{equation}
Consequently, the expected errors after $j$ target intervals are
\begin{align}
  \mathbb{E}\, e_{v,j} &= \lambda^j e_{v,0}, \\
  \mathbb{E}\, e_{u,j} &= \lambda^j e_{u,0} + j \gamma (1-\lambda) \lambda^{j-1} e_{v,0}.
  \label{eq:uniform-upstream}
\end{align}
The second term in~\eqref{eq:uniform-upstream} shows that, under uniform sampling, a stale successor error is propagated upstream and must be unlearned again after subsequent target synchronizations.
In particular, for $e_{u,0} = 0$ and $e_{v,0} = c$, the first interval yields
\begin{equation}
  \label{eq:uniform-first}
  \mathbb{E}\, e_{u,1}^{\text{uniform}} = \gamma c \bigl[1 - (1 - \eta/2)^K\bigr],
\end{equation}
which approaches the full discounted successor error $\gamma c$ as $K \to \infty$.

\subsection{Mitigation by target alignment}
\label{subsec:tarl-mitigation}

\gls{acr:tarl} can mitigate this upstream propagation.
Consider again a correct upstream value and an overestimated successor, $e_{u,0} = 0$ and $e_{v,0} = c > 0$.
Assume a batch size $m = 1$, a candidate pool that contains both transitions at every update (i.e., $b = m$ with candidates drawn without replacement), and random tie-breaking.

Because $s_1$ is terminal, $\delta_v = \bar{\delta}_v$, so $\mathcal{A}_v = 1$ as long as its TD error is nonzero.
Before the first update of $v$, we have $v = \bar{v}$ and hence $\delta_u = \bar{\delta}_u$, so both transitions have $\mathcal{A} = 1$ and each update selects either with probability $1/2$.
Once $v$ has been updated $n_v \ge 1$ times, its online value has moved towards $R$ while $\bar{v}$ remains frozen, and by~\eqref{eq:drift}
\begin{equation}
  \delta_u = \bar{\delta}_u - \gamma c \bigl[1 - (1-\eta)^{n_v}\bigr],
  \qquad
  \bar{\delta}_u = \gamma c\, (1-\eta)^{n_u} > 0,
\end{equation}
where $n_u$ is the number of preceding updates of $u$.
The online residual is therefore either smaller than the lagged residual or of opposite sign, so $\mathcal{A}_u < 1 = \mathcal{A}_v$, and $v$ is selected for the remainder of the interval.
Letting $N$ denote the number of updates of $u$ in the interval, which all occur before the first update of $v$, its probability distribution is given by
\begin{equation}
  \Pr(N = n) =
  \begin{cases}
    2^{-(n+1)}, & n = 0, \ldots, K-1, \\
    2^{-K},     & n = K.
  \end{cases}
\end{equation}
At the conclusion of the interval, the errors are
\begin{equation}
  e_{v,1} = (1-\eta)^{K-N} c, \qquad
  e_{u,1} = \gamma c \bigl[1 - (1-\eta)^{N}\bigr].
\end{equation}
Taking the expectation over $N$ yields
\begin{equation}
  \label{eq:tarl-first}
  \mathbb{E}\, e_{u,1}^{\text{TARL}}
  = \gamma c \, \frac{\eta}{1+\eta} \Bigl[1 - \Bigl(\frac{1-\eta}{2}\Bigr)^{K}\Bigr]
  \;\le\; \gamma c \, \frac{\eta}{1+\eta}.
\end{equation}
Compared with~\eqref{eq:uniform-first}, the two coincide for $K = 1$, and for $K \ge 2$ \gls{acr:tarl} propagates strictly less of the stale successor error.
As $K \to \infty$, the propagated error approaches $\gamma c$ under uniform sampling but remains bounded by $\gamma c\, \eta/(1+\eta)$ under \gls{acr:tarl}.
The benefit thus grows with the target update interval, which is consistent with the corresponding ablation in Table~\ref{tab:ablation_nauc}: a tenfold longer interval lowers the mean nAUC of \gls{acr:dqn} from $0.292$ to $0.226$, whereas that of TA-DQN changes from $0.420$ to $0.411$.

Two caveats apply.
First, the bound relies on the misaligned transition being excluded deterministically.
If candidates are drawn with replacement, $u$ is still updated with probability $1/4$ per step after the first update of $v$, and the propagated error again approaches $\gamma c$ as $K \to \infty$, although more slowly than under uniform sampling. Hence, the $\eta/(1+\eta)$ bound describes an idealized limit.
Second, the same mechanism may also defer useful updates: If $u$ carries a genuine error of its own, a drift of its successor can likewise lower $\mathcal{A}_u$, so that its correction is largely deferred until the next target synchronization, which resets the alignment of all transitions.

\section{Hardware specification}\label{sec:hardware}
The \textsc{Atari} experiments were conducted on a workstation equipped with an AMD Ryzen 9 7950X CPU (16 cores at 4.5 GHz), 128 GB of RAM, and an NVIDIA RTX 4090 GPU with 24 GB of memory (CUDA Toolkit version 12.3). All other numerical experiments were performed on a 2024 MacBook Air with an Apple M3 processor.

\section{Assets}\label{sec:assets}

We use Gymnasium \citep{gymnasium} (MIT) with MuJoCo \citep{mujoco} (Apache 2.0) and Box2D (zlib) for continuous control, the Arcade Learning Environment \citep{bellemare2013arcade} (GPL-2.0) with Atari 2600 ROMs obtained via AutoROM for Atari-10, MinAtar \citep{minatar} (GPL-3.0) for the ablations, and Stable-Baselines3 \citep{antonin_raffin_stable-baselines3_2024} and RL Baselines3 Zoo \citep{rl-zoo3} (both MIT) for the SAC implementation and its hyperparameters.


\end{document}